\newif\ifreview

\newif\ifnotes
\notestrue 

\documentclass[conference, ]{IEEEtran}
\IEEEoverridecommandlockouts
\usepackage[dvipsnames,table]{xcolor}

\usepackage{wrapfig}
\usepackage{algpseudocode}
\usepackage{nicematrix}
\usepackage{booktabs}

\usepackage[normalem]{ulem}
\usepackage{setspace}
\usepackage{siunitx} 
\usepackage{pgf}
\usepackage{tikz}
\usepackage{tikzscale}
\usetikzlibrary{arrows,automata, calc, decorations.pathreplacing, shapes,positioning}
\ifnotes
  \usepackage[nomargin,inline,draft]{fixme}
\else
  \usepackage[nomargin,inline,final]{fixme}
\fi

\usepackage{subcaption}
\usepackage[utf8]{inputenc}
\usepackage[T1]{fontenc}
\usepackage{microtype}
\usepackage{pstricks}
\usepackage{graphicx}
\usepackage{amsmath}
\usepackage{amssymb}
\usepackage{xspace}
\usepackage{listings}
\usepackage{multirow} 
\usepackage{amsthm}
\usepackage{enumitem}

\usepackage{mathtools}

\usepackage{float}
\usepackage{flushend}
\usepackage{makecell}
\usepackage{pifont}
\usepackage{xcolor}
\newcommand{\cmark}{\textcolor{green!60!black}{\ding{51}}} 
\newcommand{\xmark}{\textcolor{red}{\ding{55}}}            

\usepackage[
  sorting=none,
  style=numeric-comp,
  backend=biber,
  isbn=false,
  url=true,
  doi=false,
  url=false,
  mincrossrefs=100,
  maxnames=12,
  giveninits=true
]{biblatex}

\AtEveryBibitem{\clearname{editor}}

\definecolor{vrpink}{RGB}{255,0,127}
\definecolor{vrblue}{RGB}{30,144,255}
\definecolor{vrolive}{RGB}{85,107,47}
\definecolor{vrroyalblue}{RGB}{65,105,225}
\definecolor{brgreen}{RGB}{93,188,65}
\definecolor{ivsalmon}{RGB}{255,160,122}
\definecolor{vrlpink}{RGB}{255,192,203}
\definecolor{mvcol}{RGB}{5,150,25}
\definecolor{jdgreen}{RGB}{4,99,7}
\definecolor{jdred}{RGB}{133,6,6}

\fxusetheme{color}
\fxuseenvlayout{color}

\usepackage{graphicx}
\graphicspath{{./figures/}}
\DeclareGraphicsExtensions{.pdf,.jpeg,.png,.jpg}

\usepackage[linesnumbered,ruled,vlined]{algorithm2e}
\usepackage{algorithmicx}
\usepackage{algpseudocode}
\usepackage{algcompatible}
\algblockdefx{Event}{EndEvent}[1]
{\textbf{upon event }#1 \textbf{do }}
{\textbf{}}

\algblockdefx{Function}{EndFunction}[1]
{\textbf{Function }#1 }
{\textbf{}}

\definecolor{mymaroon}{RGB}{128,0,0}

\usepackage[
  colorlinks=true,
  linkcolor=purple,
  citecolor=purple,
  urlcolor=purple,
  pdfauthor={},
  pdftitle={},
  pdfsubject={},
  pdfkeywords={},
  bookmarks=false,
]{hyperref}

\hypersetup{colorlinks=true,urlcolor=purple,citecolor=purple,urlcolor=purple}

\usepackage{pifont}

\newcommand{\morph}{Morph}

\newcommand{\EL}{Epidemic Learning}

\newcommand{\evalInterval}{\ensuremath{\Delta_r}}





\usepackage{mathtools}
\newtagform{blue}{\color{BlueViolet}(}{)}

\DeclareUnicodeCharacter{0301}{\'{e}}

\usepackage[export]{adjustbox}

\begin{document}

\title{Dynamic Topology Optimization for Non-IID Data in Decentralized Learning}
\ifreview
  \author{
  }

  \else
  \author{\IEEEauthorblockN{Bart Cox, Antreas Ioannou, Jérémie Decouchant}
  \IEEEauthorblockA{
    \textit{Delft University of Technology}\\
    b.a.cox@tudelft.nl, a.ioannou@student.tudelft.nl, j.decouchant@tudelft.nl}
  }
 
\fi
\maketitle
\thispagestyle{plain}
\pagestyle{plain}

\begin{abstract}
Decentralized learning (DL) enables a set of nodes to train a model collaboratively without central coordination, offering benefits for privacy and scalability. However, DL struggles to train a high accuracy model when the data distribution is  non-independent and identically distributed (non-IID) and when the communication topology is static. To address these issues, we propose \morph{}, a topology optimization algorithm for DL. In \morph{}, nodes adaptively choose peers for model exchange based on maximum model dissimilarity. \morph{} maintains a fixed in-degree while dynamically reshaping the communication graph through gossip-based peer discovery and diversity-driven neighbor selection, thereby improving robustness to data heterogeneity. 
Experiments on CIFAR-10 and FEMNIST with up to $100$ nodes show that \morph{} consistently outperforms static and epidemic baselines, while closely tracking the fully connected upper bound. On CIFAR-10, \morph{} achieves a relative improvement of $1.12\times$ in test accuracy compared to the state-of-the-art baselines. On FEMNIST, \morph{} achieves an accuracy that is $1.08\times$ higher than \EL. Similar trends hold for $50$-node deployments, where \morph{} narrows the gap to the fully connected upper bound within $0.5$ percentage points on CIFAR-10. These results demonstrate that \morph{} achieves higher final accuracy, faster convergence, and more stable learning as quantified by lower inter-node variance, while requiring fewer communication rounds than baselines and no global knowledge.
\end{abstract}

\begin{IEEEkeywords}
Decentralized Learning, Heterogeneous Data Distribution, Communication Graph
\end{IEEEkeywords}

\section{Introduction}

Federated Learning (FL) has emerged as an alternative to traditional centralized machine learning, where data is aggregated in a central location, to reduce reliance on central data storage.
FL is a common distributed learning paradigm where a central coordinator orchestrates the training process by aggregating model updates from participating clients~\cite{mcmahanCommunicationEfficientLearningDeep2017, zhangSurveyFederatedLearning2021,de2024training}.
In addition, FL mitigates privacy concerns related to sensitive data being pooled on a central server~\cite{wittkoppDecentralizedFederatedLearning2021,yuProvablePrivacyAdvantages2025}, without completely eliminating them~\cite{xu2022agic,shankar2024share,mualan2024ccbnet,wang2024mudguard}.
Variants of FL have been described to support heterogeneous clients and networks, e.g., using several servers~\cite{zuo2024spyker} or asynchronous client-server interactions~\cite{cox2024asynchronous}. However, FL always requires some degree of central coordination, which can limit scalability~\cite{kairouzAdvancesOpenProblems2021, laiFedScaleBenchmarkingModel2022, lianCanDecentralizedAlgorithms2017} and create a performance bottleneck~\cite{yingBlueFogMakeDecentralized2021, maStateoftheartSurveySolving2022}. 
Decentralized Learning (DL) is a distributed learning scheme that has been proposed to eliminate the need for central coordination. In DL, nodes discover each other and communicate through peer-to-peer (P2P) or gossip-based protocols~\cite{ormandiGossipLearningLinear2013, hegedusGossipLearningDecentralized2019}. While DL mitigates many performance-related FL limitations, it also faces communication efficiency challenges. In particular, fully connected topologies are impractical in large-scale networks~\cite{kongConsensusControlDecentralized2021}, which force DL to rely on sparsely connected communication topologies.   

The communication topology used in a DL system significantly affects its communication cost, convergence rate, scalability, and final accuracy~\cite{palmieriImpactNetworkTopology2024}, especially under non-independent and identically distributed (non-IID) data conditions~\cite{gaoSemanticawareNodeSynthesis2023, barsRefinedConvergenceTopology2023,hsiehNonIIDDataQuagmire2020,cox2022aergia}, where nodes possess diverse local datasets. Many studies focused on addressing the non-IID challenge using static topologies and decentralized optimization methods such as decentralized parallel stochastic gradient descent (D-PSGD)~\cite{lianCanDecentralizedAlgorithms2017}. However, such static-topology methods often struggle to effectively handle non-IID data when the network structure lacks sufficient connectivity or exposes nodes to overly similar local data, limiting global knowledge exchange~\cite{hsiehNonIIDDataQuagmire2020}.

To overcome this, recent research explored adaptive topologies and demonstrated the benefits of dynamically adjusting the communication graph during training~\cite{linReinforcementBasedCommunication2021,devosEpidemicLearningBoosting2023,menegattiDynamicTopologyOptimization2024}. However, many such methods require some form of global knowledge or lack mechanisms for dynamic adaptation, limiting their scalability and robustness in heterogeneous settings. It is therefore still an open issue to design a fully decentralized approach that explicitly accounts for non-IID data while enabling intelligent dynamic peer selection (as shown in Table~\ref{tab:method-comparison}).

We introduce a fully decentralized method, named \morph{}, 
that enables nodes to select their neighbors based on local model dissimilarity, without relying on any form of global knowledge or central orchestration. Each node dynamically evaluates and adjusts its incoming connections from which it receives others' models to update its own.
Additionally, \morph{} enables nodes to progressively discover new peers over time, expanding their local view of the network and their optimization opportunities using indirect dissimilarity estimation. 

As a summary, this work makes the following contributions: 

$\bullet$ We propose \morph{}, a novel fully decentralized framework that dynamically adjusts the communication topology based on local model dissimilarity. \morph{} allows nodes to optimize their incoming connections without global information or centralized coordination. 
\morph{} maintains a fixed in-degree per node by probabilistically selecting diverse peers for incoming, rather than outgoing, connections. This guarantees that every node is exposed to external information in every round, mitigating local overfitting under non-IID data.
To enable peer discovery, nodes exchange information about their known neighbors during model updates, progressively expanding their local view of the network.

$\bullet$ We describe methods that allow nodes to optimize their incoming connections in decentralized systems. To identify the nodes whose model they should receive, \morph{} nodes first evaluate the dissimilarity between their local models and those they received using cosine similarity. \morph{} further allows nodes to infer model dissimilarity with unknown peers via gossip, enabling informed peer selection even under partial network knowledge. This enhances adaptability in sparse and evolving topologies. Nodes then update their neighborhood probabilistically based on softmax sampling to select the nodes whose models differ the most from theirs while avoiding redundancy among incoming models.    

$\bullet$ We evaluate \morph{} on the CIFAR-10~\cite{krizhevsky2009learning} and FEMNIST~\cite{caldasLEAFBenchmarkFederated2019} datasets under realistic non-IID settings. As shown in Table~\ref{tab:decentralized-accuracy}, \morph{} achieves consistently higher accuracy than static and epidemic baselines, while closely tracking the fully connected upper bound. On CIFAR-10, \morph{} improves the accuracy by $1.13\times$ compared to the baselines. On FEMNIST, \morph{} is up to $1.08\times$ better than the baselines. Across both datasets and node counts, \morph{} consistently closes the gap to the fully connected baseline while offering improved robustness and efficiency.

\section{Background}
\subsection{System Model}

We consider a decentralized learning (DL) system \(\mathcal{N}\) that consists of a set of distributed computational nodes \( \{1, 2, \dots, n\} \), which collaborate to train a model. 
Each node \( i \in [1,n] \) holds a private dataset that follows a distribution \( \mathcal{D}^{(i)} \), which may differ from the one of other nodes, over a data space \( \mathcal{Z} \), and on which it can perform computations. 

Communication among nodes occurs over a network topology represented by a directed graph \( G = (V, E) \), where each node corresponds to a vertex \( v \in V \), and an edge \( (j, i) \in E \) indicates that node \( j \) can send information directly to node \( i \). 
This communication model is inspired by classical peer-to-peer (P2P) systems, in which nodes operate as equal participants, both consuming and supplying information~\cite{schollmeierDefinitionPeertopeerNetworking2001, engkeongluaSurveyComparisonPeertopeer2005}. In such systems, a P2P peer discovery service periodically provides each node with a set of new potential neighbors, enabling continuous exploration of the network. Randomized gossip protocols are often used to propagate information efficiently without centralized scheduling~\cite{mokhtar2014acting,decouchant2016pag,kempeGossipbasedComputationAggregate2003}. In our settings, for simplicity, we assume that nodes know their neighbors in an initial graph and learn about other nodes by exchanging information with their neighbors. 

Nodes also use the communication graph to train a model by exchanging model updates with their neighbors. Connections between nodes may evolve over time, following our topology adaptation mechanisms. The out-degree of a node is the number of other nodes it transmits information to, while its in-degree is the number of nodes from which it receives information.
We assume that the initial communication graph is connected in the undirected sense, that is, if edge directions are ignored, there exists a path between any pair of nodes. While each node initially communicates only with a subset of neighbors, we assume that nodes can, in principle, establish connections with any other node, provided they are aware of its existence (e.g., via the P2P discovery service). 

{
\SetAlgoLined
\DontPrintSemicolon
\begin{algorithm}[t]
\caption{Epidemic Learning}
\label{alg:epidemic_learning}
\small

\textbf{Require}: Initial model $x_0^{(i)} = x_0 \in \mathbb{R}^d$, number of round $T$, step-size $\gamma$, sample size $k$.\; 
\For{$t=0,\dots,T-1$}{
Randomly sample a data point $\xi_t^{(i)}$ from the local data distribution $\mathcal{D}^{(i)}$\;
Compute the stochastic gradient $g_t^{(i)} := \nabla f(x_t^{(i)}, \xi_t^{(i)})$\;
Partially update local model $x_{t+\frac{1}{2}}^{(i)} := x_t^{(i)} - \gamma g_t^{(i)}$ \tcp{Line 6-9: Random communication phase}
Sample $k$ other nodes from $[n] \setminus \{i\}$ using EL-Oracle or EL-Local\;
Send $x_{t+\frac{1}{2}}^{(i)}$to the selected nodes\;
Wait for the set of updated models $S_t^{(i)}$ \tcp{$S_t^{(i)}$ is the set of received models by node $i$ in round $t$}
Update $x_t^{(i)}$ to the average of available updated models according to (2)\;
}

\end{algorithm}
}

\subsection{Decentralized Learning}

We consider the standard decentralized learning objective in which a group of \( n \) nodes seeks to collaboratively minimize a global loss function by performing local updates and exchanging information with neighbors.  Let \( f: \mathbb{R}^d \times \mathcal{Z} \rightarrow \mathbb{R} \) be a loss function that evaluates model performance on a data point. The local loss function at node \( i \) is defined as the expectation over its local distribution:
\begin{equation}
f^{(i)}(x) := \mathbb{E}_{\xi \sim \mathcal{D}^{(i)}} [f(x, \xi)].
\end{equation}

The goal of the decentralized learning system is to minimize the average loss over all nodes:
\begin{equation}
\min_{x \in \mathbb{R}^d} F(x) := \frac{1}{n} \sum_{i=1}^n f^{(i)}(x).
\end{equation}

A classical decentralized learning algorithm follows  Algorithm~\ref{alg:epidemic_learning} and proceeds in synchronous rounds. In each round,  a node $i$ first trains its model on its local data. It then selects $k$ nodes in the network to which it will send its updated model. Similarly, node $i$ receives the model of some other nodes that connected to it, and, at the end of a training round, sets its model to the average of the received models. 

\begin{figure*}[t]
    \centering
    \includegraphics[width=1.0\linewidth]{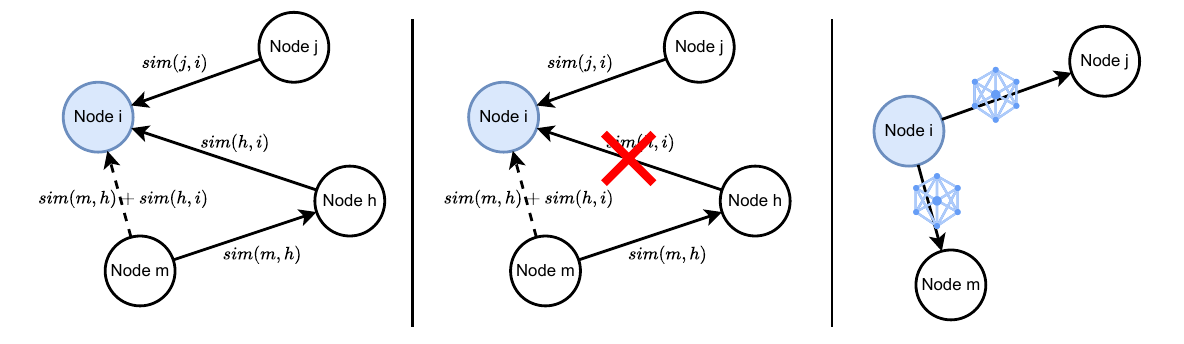}
    \caption{Node $i$ gets connection requests from three requesting nodes $j$, $h$ and $m$ that share their dissimilarity value with $i$. Node $m$ had approximated its dissimilarity with node $i$ using the cosine inequality. Node $i$ select the top-k connection requests (here $k=2$) and uses its new outgoing connections to share its model updates. 
}
    \label{fig:sim_selection}
\end{figure*}

\section{\morph{}}
\morph{} is based on a fully decentralized topology adaptation mechanism that dynamically updates each node’s communication neighborhood based on model dissimilarity. \morph{} aims at letting nodes receive models that differ from theirs as much as possible, while keeping the communication graph connected so that the model of all nodes converge similarly.  

\subsection{Evaluating Peer Diversity}
\label{topology_adaptation}

In \morph{}, nodes receive models directly from their incoming connections and can therefore directly evaluate their dissimilarity with them. However, they also require a way to evaluate their dissimilarity with other nodes. We explain in this section how \morph{} uses cosine similarity for this purpose.

To quantify model diversity, we compute the cosine similarity between a node's local model \( w_i \) and a candidate peer's model \( w_j \). To avoid domination by large layers, similarity is computed per layer and averaged across layers. Denoting parameters of layer \( l \) by \( \theta_l^{(i)} \) and \( \theta_l^{(j)} \), we define
\begin{equation}
\begin{split}
       \text{sim}(w_i, w_j) &= \frac{1}{L} \sum_{l=1}^{L} \text{sim}_l, \\
       \text{where } \text{sim}_l &= \frac{\theta_l^{(i)} \cdot \theta_l^{(j)}}{\|\theta_l^{(i)}\|_2 \cdot \|\theta_l^{(j)}\|_2}.
\end{split}
\end{equation}
Cosine similarity is invariant to parameter scaling, efficient to compute, and incurs minimal communication overhead~\cite{zecEffectsSimilarityMetrics2024}.  

When direct access to a peer's model is unavailable, similarity is estimated via transitive inference. Suppose node \( i \) has both the model of an intermediate peer \( y \) and a reported similarity between \( y \) and a target peer \( z \). Then, the estimate is
\begin{equation}
    \hat{\text{sim}}(w_i, w_z) = \frac{1}{|\mathcal{H}_z|} \sum_{(t,y,\sigma_{yz}) \in \mathcal{H}_z} \text{sim}(w_i, w_y) \cdot \sigma_{yz},
\end{equation}
where \( \mathcal{H}_z \) stores the five most recent similarity reports for peer \( z \). Although cosine similarity is not strictly transitive, the angular inequality~\cite{schubertTriangleInequalityCosine2021}:
\begin{align*}
    \arccos(\text{sim}(w_i, w_k)) \leq &\arccos(\text{sim}(w_i, w_j))\\& + \arccos(\text{sim}(w_j, w_k)),
\end{align*}
provides a theoretical bound, and empirical results show that quasi-transitive reasoning improves peer selection under noise~\cite{arandjelovicLearntQuasiTransitiveSimilarity2016}. 

\subsection{Negotiating Incoming and Outgoing Connections}

At a high level, in each round \(t\), every node \(i\) in \morph{} executes Algorithm~\ref{alg:adaptive-dpsgd}. The procedure is governed by two parameters: \evalInterval, which controls how frequently a node updates its neighbor set, and \(\beta\), which determines the stochasticity of neighbor selection via a softmax distribution over model similarities (see Figure~\ref{fig:sim_selection}). After completing local training (Alg.~\ref{alg:adaptive-dpsgd}, l.~\ref{l:training}), if the current round \(t\) is a multiple of \evalInterval, node \(i\) updates its preferred neighbors (\texttt{UpdateWantedSenders}, l.~\ref{l:updateWanted}) and issues or withdraws connection requests accordingly. It then establishes incoming connections with a set of nodes \(\mathcal{V}\) (l.~\ref{l:sendRequests}), handles outgoing connections (l.~\ref{l:receiveRequests}), sends its model to outgoing peers along with its similarity with other nodes (l.~\ref{l:sendModels}), and receives models and similarity values from incoming ones (l.~\ref{l:receiveModels}), along with limited metadata such as peer lists for neighbor discovery. Finally, node \(i\) aggregates all received models with its own using uniform averaging (l.~\ref{l:aggregateModels}). At this stage node \(i\) also updates its similarity with other nodes, possibly indirectly using the cosine angular inequality. 

Unlike in traditional decentralized learning algorithms (e.g., Alg.~\ref{alg:epidemic_learning}) where nodes send their updates to some random nodes (i.e., push-based), \morph{} involves negotiations that allow each node to decide the nodes it receives updates from (i.e., pull-based) and the nodes to which it sends its updates to. 

Once a node has computed its dissimilarity, directly or indirectly, with other nodes, it computes its new candidate set $\mathcal{C}_b$ of $k$ neighbors. This set is initially empty, and grows iteratively following a stochastic procedure, which favors diversity. During this iterative process, a node $j$ in the set of potential neighbors $\mathcal{C}_A$ is selected with probability 
\begin{equation}
    p_j = \frac{\exp\!\big(-\beta \cdot \mathrm{sim}(w, w_j)\big)}
    {\displaystyle\sum_{i \in \mathcal{C}_A \setminus \mathcal{C}_b} 
    \exp\!\big(-\beta \cdot \mathrm{sim}(w, w_i)\big)}, 
    \qquad j \in \mathcal{C}_A \setminus \mathcal{C}_b,
    \label{eq:biased_sampling}
\end{equation}
where $\beta > 0$ controls distribution sharpness. 
Nodes sample $k$ peers sequentially upon a successful connection request, i.e., $j_t \sim p_j$, updating $\mathcal{C}_b \leftarrow \mathcal{C}_b \cup \{j_t\}$. The use of a softmax function allows selecting the most dissimilar nodes with a greater priority than others. 

We now detail the phases that lines~\ref{l:sendRequests} and~\ref{l:receiveRequests} of Alg.~\ref{alg:adaptive-dpsgd} encompass. 
\morph{} keeps every node’s in-degree bounded and constant, avoiding both isolation and overfitting, while preserving diversity in received models. To further balance connectivity, \morph{} attempts to  impose an out-degree cap: each node aims at sending its model to at most $k$ other nodes that contact it. We solve this problem in a way that is analogous to the classical college admission problem~\cite{shapelyGaleS13}. Upon receiving a connection request, a node accepts it if it has less than $k$ outgoing connections. If not, it accepts it if this connection request has a greater dissimilarity than one it already accepted. Nodes whose connection is rejected, canceled, or accepted are informed, and might have to look for another connection to maintain $k$ outgoing connections. This matching always terminates in at most $\lceil (n-1)/k \rceil$ steps. Given the duration of a training round, the neighbor identification process fits  within a training round and is executed concurrently with it.

\begin{algorithm}[t]
\caption{\morph{}'s learning algorithm at node $i$}
\label{alg:adaptive-dpsgd}
\small
\KwIn{Local model $w_0$, initial neighbors $\mathcal{N}_i$, total rounds $T$, evaluation frequency \evalInterval}
\textbf{Initialization:} Set known peers $\mathcal{P}_i \leftarrow \mathcal{N}_i$\;
Wanted Senders: $w_s$ $\leftarrow$ outgoing neighbors in $\mathcal{N}_i$\;
\For{$t \leftarrow 1$ \KwTo $T$}{
    $x^{(i)}_{t+1/2} \gets x^{(i)}_t \gamma\nabla F \left( x^{(i)}_t, \xi^{(i)}_t \right)$ \; \label{l:training}
    \If{$t \bmod \evalInterval \equiv 0$} {
        $w_s \gets \texttt{UpdateWantedSenders()}$\; \label{l:updateWanted}
    }
    Request models from $\forall p \in w_s$\; \label{l:sendRequests}
    Receive requests $w_r$ from peers\; \label{l:receiveRequests}
    Send $x^{(i)}_{t+1/2}$ to $\forall p \in w_r$\; \label{l:sendModels}
    Wait for the set of updated models $S^{(i)}_t$ from $w_s$ \; \label{l:receiveModels}
    Update $\mathcal{P}_i$ using new peer information received from $w_s$\; \label{l:receivePeers}
    $x^{(i)}_{t+1} \gets \frac{1}{|S^{(i)}_t| + 1}\left(x^{(i)}_{t+1/2} + \sum_{j\in S^{(i)}_t}x^{(j)}_{t+1/2} \right)$ \label{l:aggregateModels}
}
\end{algorithm}

\subsection{Connected Topology through Random Neighbor Selection} 

\begin{algorithm}[t]
\caption{\texttt{UpdateWantedSenders} at node $i$}
\label{alg:update-wanted-senders}
\KwIn{Local model $w$, local candidate set $\mathcal{C}_A$, full candidate set $\mathcal{C}$, view size $s$, temperature $\beta$, number of biased selections $k$}
\KwOut{Partial view $\mathcal{V}$ of size $s$}

Initialize $\mathcal{C}_b \leftarrow \emptyset$
\For{$t = 1$ \KwTo $k$}{
  Compute softmax weights over remaining candidates in $\mathcal{C}_A \setminus \mathcal{C}_b$:
  \[
    p_j = \frac{\exp(-\beta \cdot \mathrm{sim}(w, w_j))}
    {\sum_{i \in \mathcal{C}_A \setminus \mathcal{C}_b} \exp(-\beta \cdot \mathrm{sim}(w, w_i))},\, j \in \mathcal{C}_A \setminus \mathcal{C}_b 
  \]
  Sample $j_t \sim p_j$ and update $\mathcal{C}_b \leftarrow \mathcal{C}_b \cup \{j_t\}$
}

Let $\mathcal{R}$ be a uniform random sample of size $s-k$ from $\mathcal{C} \setminus \mathcal{C}_A$

$\mathcal{V} \leftarrow \mathcal{C}_b \cup \mathcal{R}$\;
\Return{$\mathcal{V}$}
\end{algorithm}

While similarity-driven selection aims at accelerating convergence, it risks fragmenting the network into tightly connected clusters that block global information flow. In decentralized learning this fragmentation harms convergence, robustness, and fairness: distant regions of the population may never exchange useful updates. To prevent this, neighbor selection must balance two goals—retaining diversity for efficiency while ensuring connectivity for global mixing.

To mitigate the risk of segmentation, we use a two-step peer-sampling protocol, which has been shown to ensure biased neighborhood and a connected graph~\cite{BrahmsBortnikovGKKS08}. We first construct a biased candidate set and then performs secure re-sampling to produce near-uniform peer selections resilient to adversarial bias. In our design, the biased step corresponds to similarity-based sampling (Eq.~\ref{eq:biased_sampling}), while the unbiased step periodically injects a random set $\mathcal{R}$ of peers. These random edges reconnect clusters, ensure fairness, and provide resilience against Byzantine sampling attacks~\cite{BrahmsBortnikovGKKS08}. 

\begin{figure}
    \centering
    \includegraphics[width=1.0\linewidth]{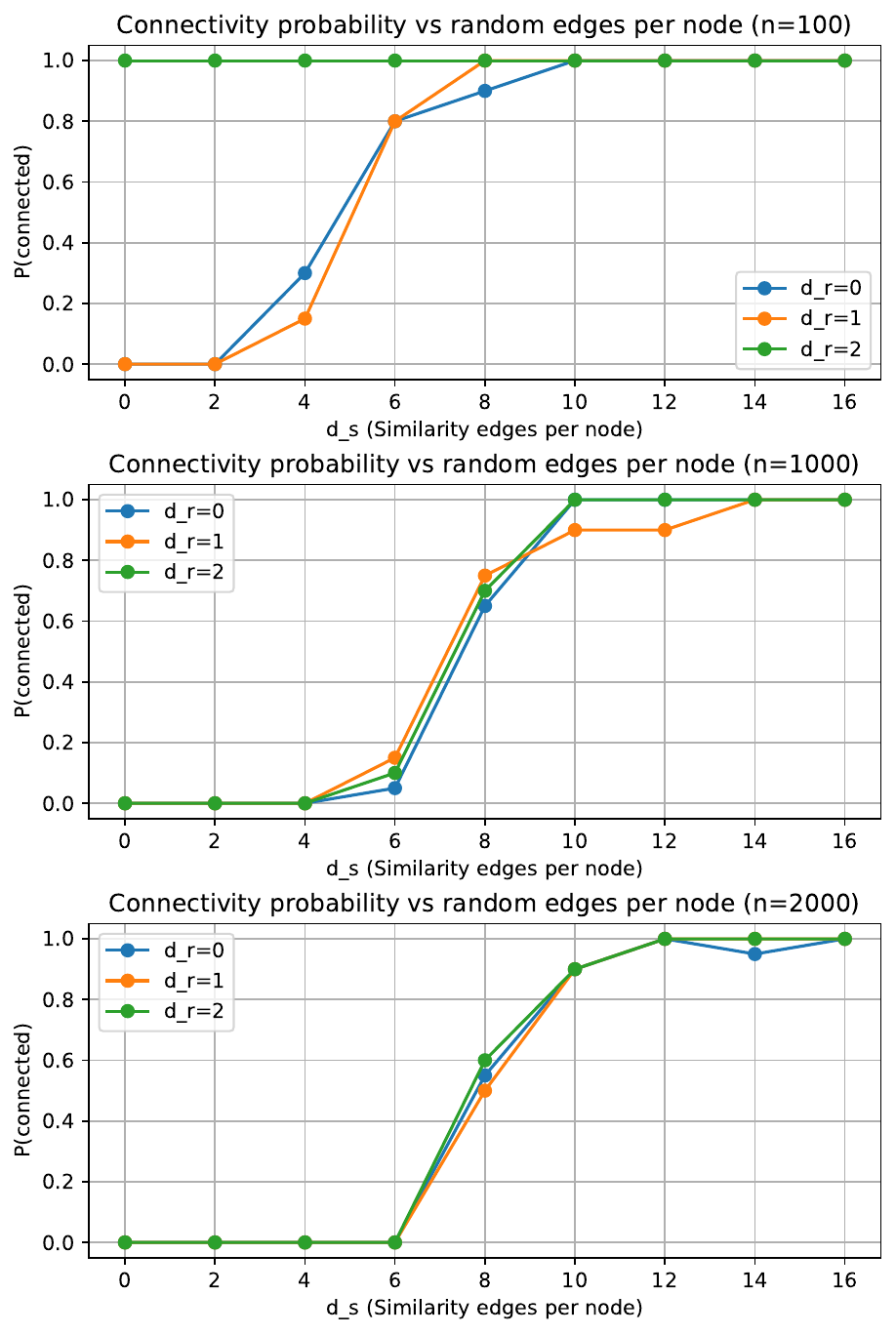}
    \caption{Probability for the communication graph to be connected depending on the number $d_s$ of connections selected using peer dissimilarity and the number $d_r$ of connections selected randomly  with different system sizes ($n=100, 1000, 2000$). In experiments, one has to choose $d_r$ and $d_s$ values that minimize $d_r+d_s$ and such that the communication graph is always connected.}
    \label{fig:connectivity_sim}
\end{figure}

Concretely, each node augments its similarity-based selection $\mathcal{C}_b$ with a uniformly random sample $\mathcal{R} \subseteq \mathcal{C} \setminus \mathcal{C}_A$ of size $s-k$. The final neighborhood is
\begin{equation}
    \mathcal{V} = \mathcal{C}_b \cup \mathcal{R}.
\end{equation}
This hybrid design, both similarity-based and random-based, combines the strengths of both approaches: similarity edges accelerate local adaptation, while random (re-sampled) edges maintain global connectivity. The added overhead is only $O(\log n)$ messages per node per round, with mixing-time overhead also $O(\log n)$, ensuring scalability in practice. Simulations (Figure~\ref{fig:connectivity_sim}) confirm that even a small random set $\mathcal{R}$ (two peers per node) suffices to prevent network segmentation.

\section{Evaluation}
\label{evaluation}

\begin{figure*}[t]
  \centering
    \begin{subfigure}[t]{0.32\textwidth}
        \centering
        \includegraphics[width=\linewidth]{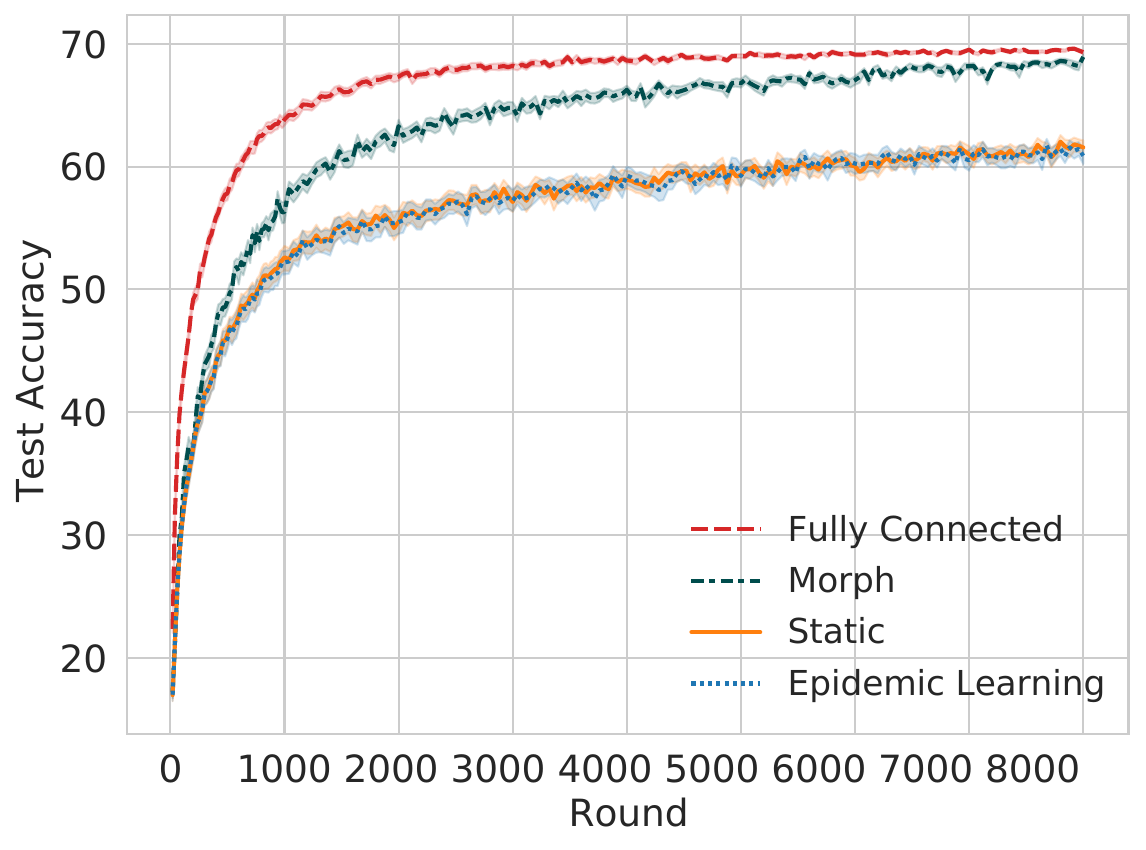}
        \caption{Test Accuracy}
    \end{subfigure}
    ~ 
    \begin{subfigure}[t]{0.32\textwidth}
        \centering
        \includegraphics[width=\linewidth]{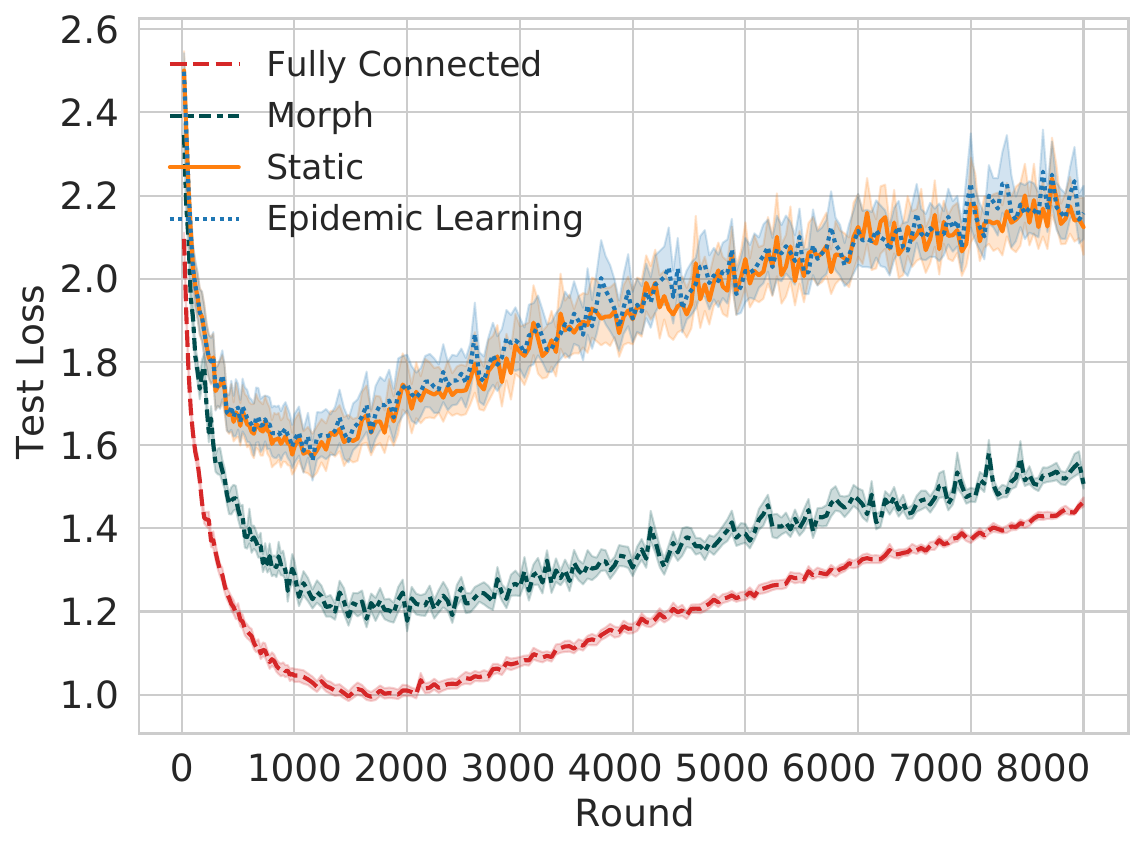}
        \caption{Test Loss}
    \end{subfigure}
    \begin{subfigure}[t]{0.32\textwidth}
        \centering
        \includegraphics[width=\linewidth]{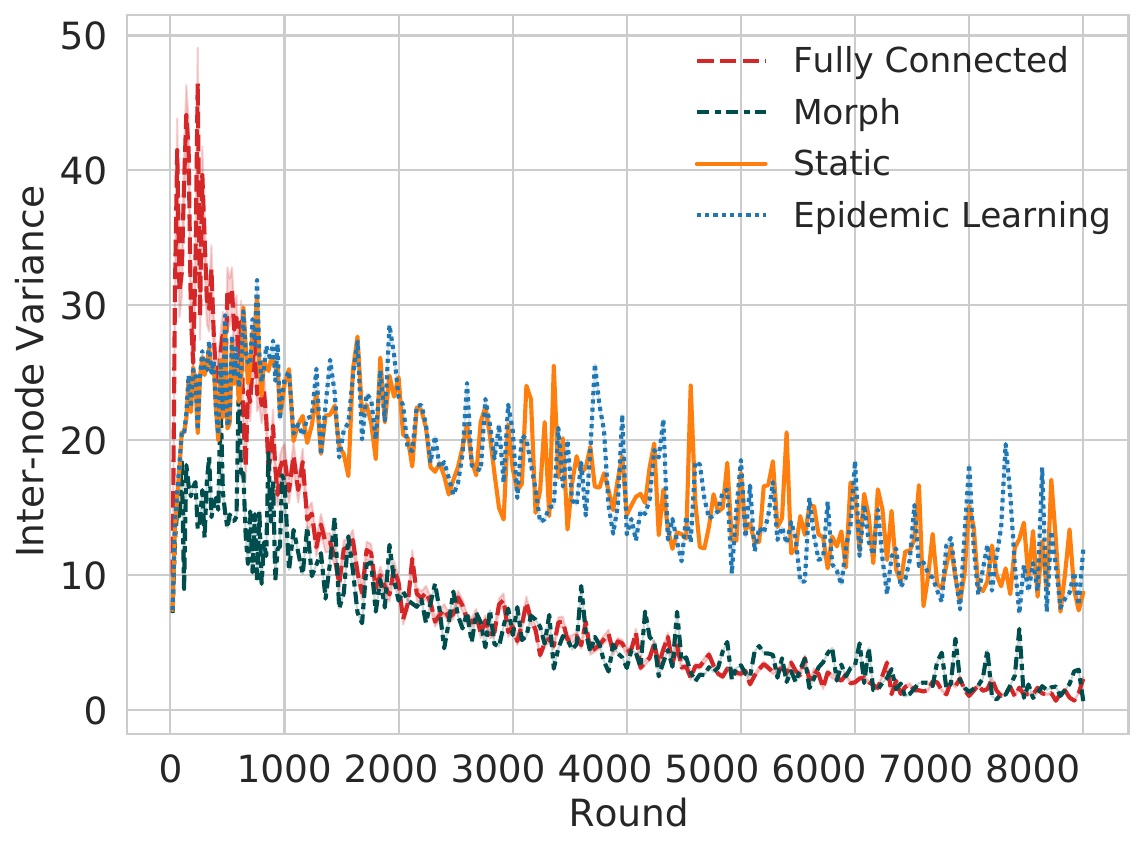}
        \caption{Inter node variance}
        \label{fig:inv_general}
    \end{subfigure}
  \caption{
Performance comparison on CIFAR-10 with 100 nodes in a non-IID setting using degree-3 topologies. 
The panels show: \textbf{(a)} mean top-1 test accuracy over communication rounds (shaded regions denote standard deviation across five runs), 
\textbf{(b)} mean test loss, and 
\textbf{(c)} inter-node variance, i.e., the variance of per-node test accuracies across the entire system. 
Inter-node variance captures fairness and consistency: lower values indicate that nodes converge to similar performance levels. 
Epidemic Learning (EL) suffers from high inter-node variance ($\approx 15.5$), reflecting severe inconsistency across nodes, 
while \morph{} matches the stability of the fully connected topology (variance $<0.02$) at far lower communication cost. 
}
  \label{fig:general_accuracy}
\end{figure*}

\subsection{Experimental Setup}

\subsubsection{Datasets and Partitioning}

\textbf{CIFAR-10} is a widely-used image classification dataset consisting of 60{,}000 $32\times32$ color images across 10 classes~\cite{krizhevsky2009learning}. To simulate a non-IID data distribution, we partition the dataset across clients using a Dirichlet distribution~\cite{hsuMeasuringEffectsNonIdentical2019} with a concentration parameter $\alpha = 0.1$. This results in each client having a different class distribution.

\textbf{FEMNIST} is a federated version of the Extended MNIST dataset, containing handwritten characters from 62 classes written by 3,550 users~\cite{caldasLEAFBenchmarkFederated2019}. 

\subsubsection{Implementation}
Our implementation of \morph{}\footnote{Code available at: \url{https://github.com/bacox/Morph}} builds on the decentralized parallel SGD (D-PSGD) framework provided by the DecentralizePy library~\cite{dhasadeDecentralizedLearningMade2023}. We extend this framework to incorporate \morph{}'s dissimilarity-guided neighbor selection. The communication topology is initialized as either a random 100-node 7-regular or 3-regular graph, and is dynamically updated during training. Specifically, the topology is re-evaluated every $\evalInterval = 5$ communication rounds to account for evolving contribution dynamics, using a softmax temperature of $\beta = 500$.

Experiments are conducted in Python~3.11.2 on two servers with 64-core processors (2 threads per core) and 500~GB memory, without GPUs. A decentralized system is emulated using 100 parallel processes, each representing a network node, with shared CPU and memory resources. Each run spans 8,000 communication iterations, and all pseudo-random generators use a fixed seed for reproducibility.
For CIFAR-10, we evaluate two 100-node communication graphs across five independent runs with different seeds. The first graph has degree 7, while the second has degree 3, approximating the connectivity bound $\mathcal{O}(\log n)$ for $n$ nodes.

\subsubsection{Baselines}
We benchmark \morph{} against three representative decentralized learning baselines, all derived from variants of D-PSGD.
\begin{itemize}
    \item \texttt{Static}, which employs a static 3-regular or 7-regular random graph, consistent with the initial topology in our method, and uses the Metropolis-Hastings (MH) averaging scheme to mitigate topological bias.  
    \item \texttt{Fully connected}, which adopts a fully connected topology, representing an optimistic upper bound on achievable performance.  
    \item \texttt{Epidemic Learning}~\cite{devosEpidemicLearningBoosting2023}, which samples a random $k$-regular topology at each communication round; we set $k \in \{3,7\}$ to align the communication volume with our implementation.
\end{itemize}

\subsubsection{Evaluation Metrics}
We evaluate performance using four metrics: mean accuracy, mean test loss, inter-node variance, and total communication cost. All results are averaged over five independent runs with different seeds and reported across communication rounds.

Mean accuracy and test loss are computed by evaluating each node's model on a shared test set every 20 rounds until round 1,000 and every 40 rounds thereafter, then averaging across all 100 nodes. Test loss is measured using cross-entropy. Inter-node variance, which captures stability, is computed at the same evaluation rounds by measuring the variance of test accuracies across nodes, averaged across the five runs. 
Beyond tracking these metrics over time, we also assess communication and convergence efficiency by measuring the number of rounds and the volume of communication required for each method to reach the best accuracy (achieved by \EL).

\begin{figure*}[t]
  \centering
    \begin{subfigure}[t]{0.32\textwidth}
        \centering
        \includegraphics[width=\linewidth]{figures/FIG3.pdf}
        \caption{Connectivity $=3$}
    \end{subfigure}
    ~ 
    \begin{subfigure}[t]{0.32\textwidth}
        \centering
        \includegraphics[width=\linewidth]{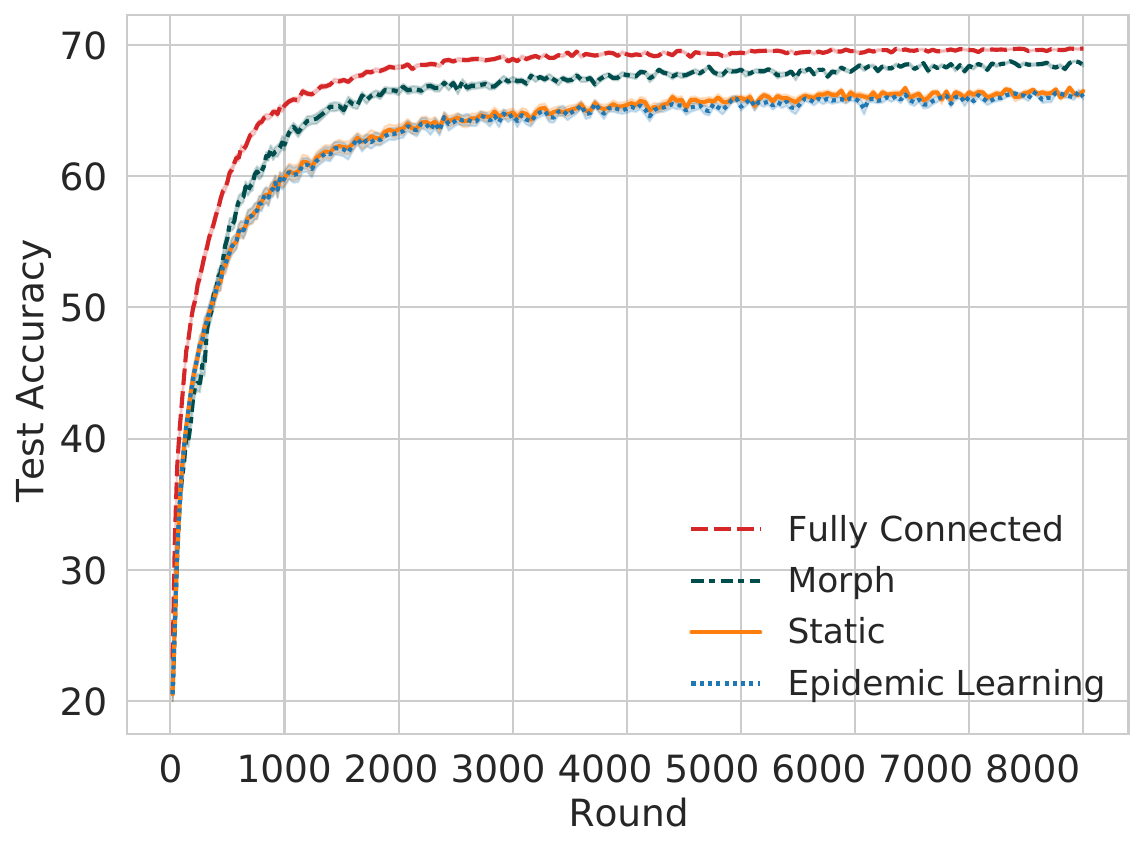}
        \caption{Connectivity $=7$}
    \end{subfigure}
    \begin{subfigure}[t]{0.32\textwidth}
        \centering
        \includegraphics[width=\linewidth]{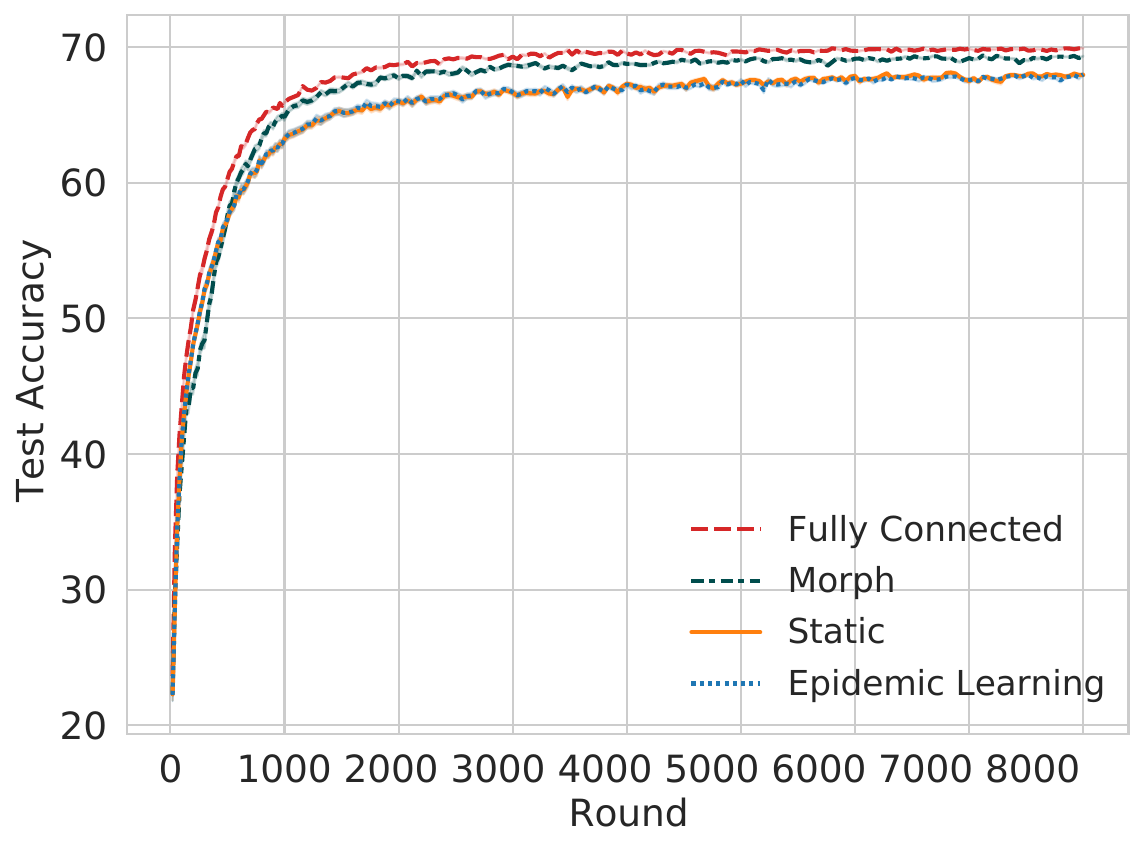}
        \caption{Connectivity $=14$}
    \end{subfigure}
  
\caption{
Test accuracy on CIFAR-10 with $100$ nodes under different connectivity levels ($k=3,7,14$). 
\morph{} consistently approaches the performance of the fully connected topology across all connectivities, while Epidemic Learning lags behind, especially at low connectivity. The Static topology reaches competitive accuracy only at $k=7$, but is less stable across other settings. Higher connectivity reduces the performance gap between methods, with \morph{} maintaining accuracy close to the upper bound. 
}
  \label{fig:cifar_abl_k}
\end{figure*}

\begin{figure*}[t]
  \centering
    \begin{subfigure}[t]{0.49\linewidth}
        \centering
        \includegraphics[width=\linewidth]{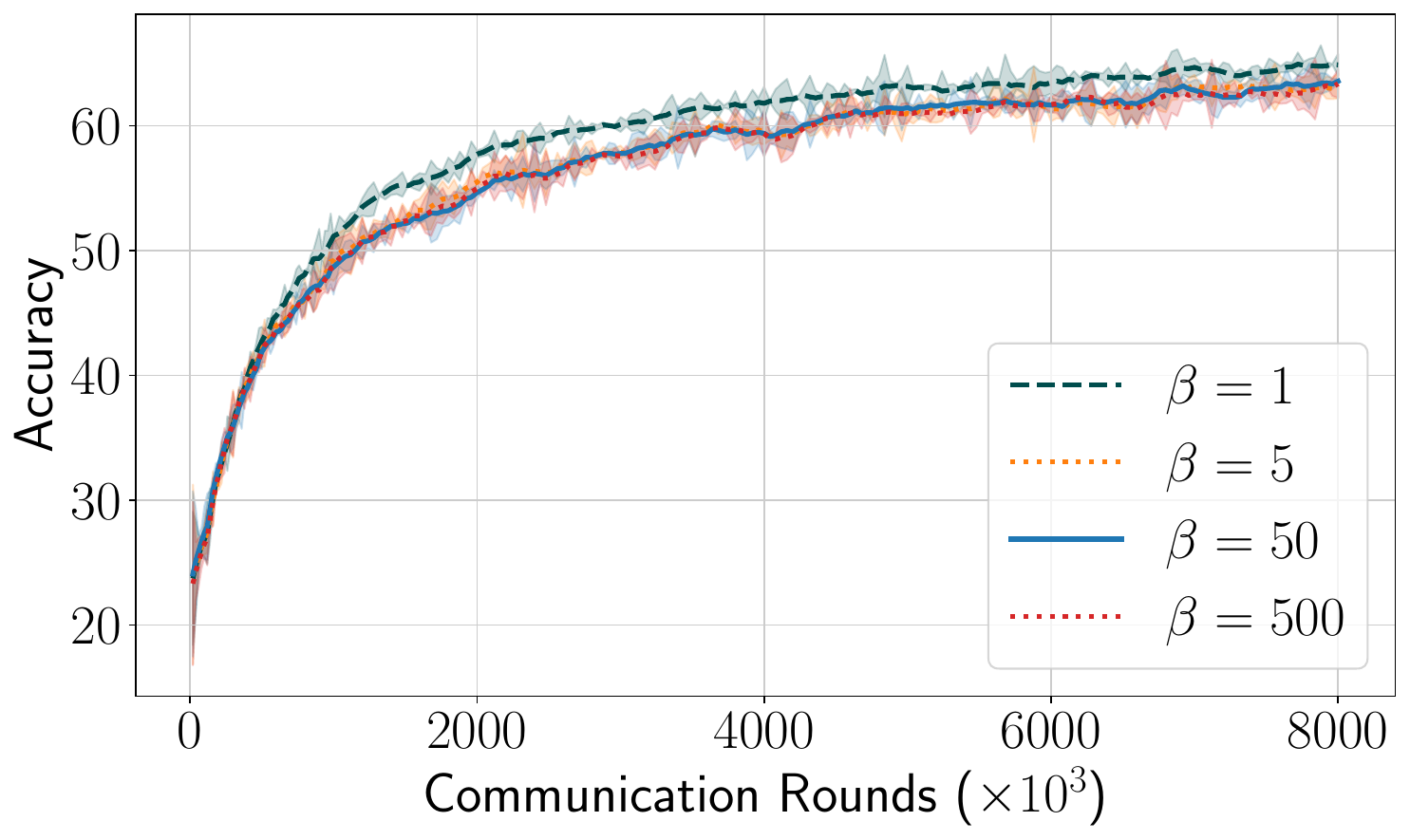}
    \end{subfigure}
    \hfill
    \begin{subfigure}[t]{0.49\linewidth}
        \centering
        \includegraphics[width=\linewidth]{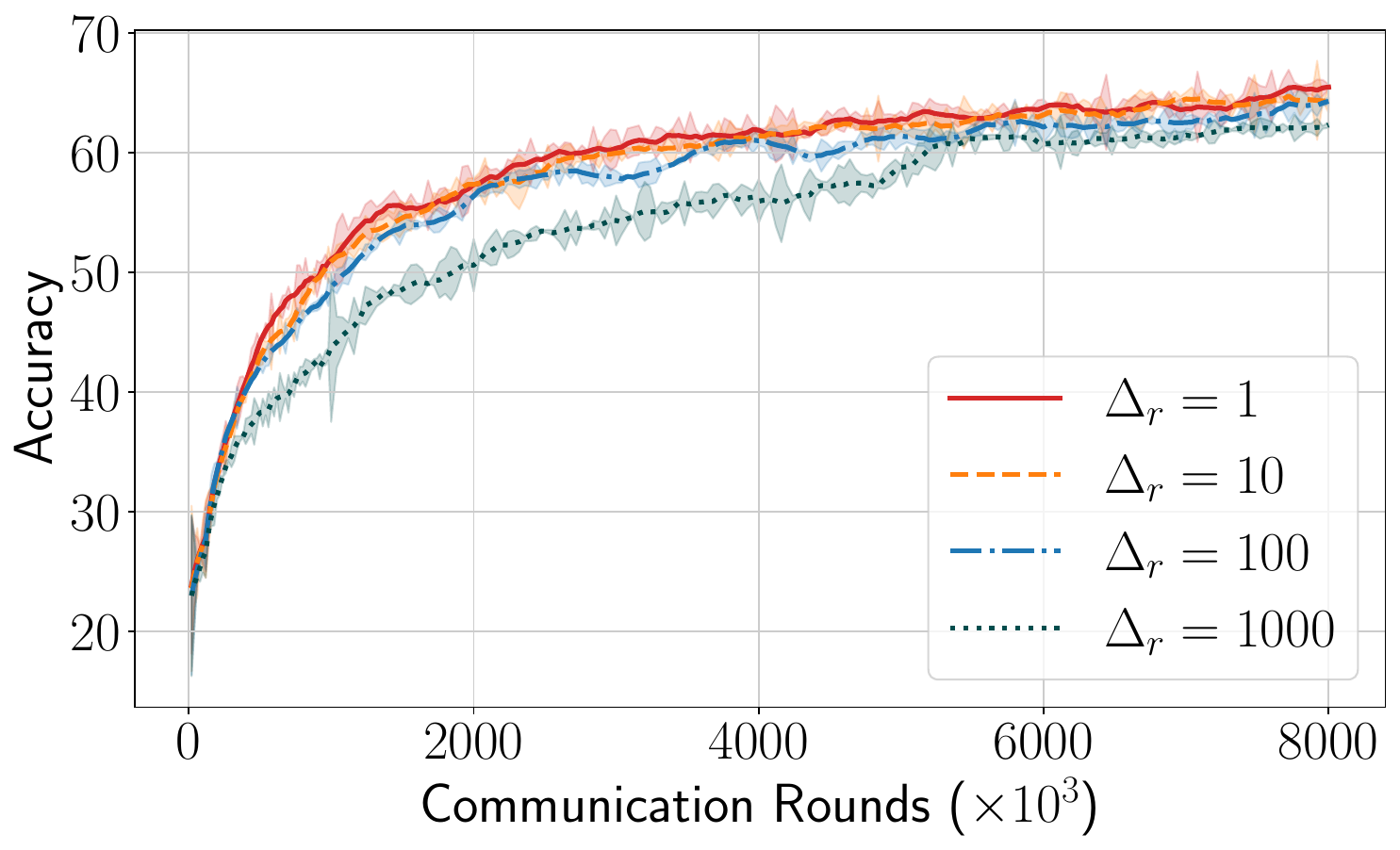}
    \end{subfigure}
  \caption{
    Ablation study on the effect of hyperparameters in \morph{} using CIFAR-10 with $100$ nodes. \textbf{Left:} Impact of the softmax sharpness parameter $\beta$. \textbf{Right:} Impact of the similarity evaluation interval \evalInterval. Lower $\beta$ improves learning performance, while values of $\Delta_r < 1000$ have little influence on convergence speed.
  }
  \label{fig:abl_beta_sim_interval}
\end{figure*}

\begin{table*}[ht]
\centering
\caption{Accuracy values on FEMNIST and CIFAR-10 with 50 and 100 nodes.}
\label{tab:decentralized-accuracy}
\begin{tabular}{lcccc}
\toprule
\multirow{2}{*}{Algorithm} & \multicolumn{2}{c}{FEMNIST} & \multicolumn{2}{c}{CIFAR-10} \\
\cmidrule(lr){2-3} \cmidrule(lr){4-5}
 & 50 nodes & 100 nodes & 50 nodes & 100 nodes \\
\midrule
Fully Connected & $64.5 \pm 1.8$ & $62.0 \pm 1.9$ & $69.5 \pm 1.5$ & $69.3 \pm 1.8$ \\ 
\midrule
Static & $57.5 \pm 2.3$ & $55.5 \pm 2.4$ & $62.5 \pm 2.1$ & $61.5 \pm 2.5$ \\
Epidemic Learning~\cite{devosEpidemicLearningBoosting2023} & $59.0 \pm 2.2$ & $57.4 \pm 2.8$ & $64.1 \pm 2.1$ & $60.8 \pm 2.2$ \\ 
\morph{} (ours) & $62.0 \pm 2.0$ & $60.0 \pm 2.5$ & $69.0 \pm 1.7$ & $68.9 \pm 2.2$ \\ 
\bottomrule
\end{tabular}
\end{table*}

\subsection{Learning Accuracy}
Our first set of experiments considers the CIFAR-10 dataset under decentralized topologies of degree three, except for the fully connected configuration which serves as an upper-bound baseline. Unless otherwise noted, we primarily discuss the 100-node setting, while Table~\ref{tab:decentralized-accuracy} provides a detailed comparison across both 50-node and 100-node scenarios. The results are visualized in Figure~\ref{fig:general_accuracy}. 

As expected, the fully connected topology consistently provides the highest accuracy, achieving $69.3\%$ on CIFAR-10 with 100 nodes. However, this comes at the cost of more than twice the communication overhead compared to sparse topologies. Our proposed method, \morph{}, achieves nearly the same performance ($68.9\%$), while requiring significantly fewer communication rounds and overall communication cost. Specifically, \morph{} achieves a $1.12\times$ higher accuracy to the best top-1 accuracy obtained by \EL. The static Metropolis-Hastings-based topology performs the worst, plateauing at $61.5\%$, more than $7$ percentage points below our method.

In the 50-node CIFAR-10 experiments, we observe a similar trend. The fully connected baseline reaches $69.5\%$, while \morph{} closely follows at $69.0\%$, clearly outperforming both \EL{} ($64.1\%$) and the static topology ($62.5\%$). These results confirm that our approach scales favorably with network size, preserving competitive accuracy even under reduced connectivity.

Turning to FEMNIST, we find that the relative advantages of \morph{} persist across both scales. With 100 nodes, the fully connected configuration again sets the upper bound at $62.0\%$. Our method achieves $60.0\%$, outperforming \EL{} ($57.4\%$) and the static topology ($55.5\%$) by margins of $2.6$ and $4.5$ percentage points, respectively. Importantly, in the 50-node case, \morph{} obtains $62.0\%$, essentially matching the fully connected network ($64.5\%$) within statistical variation, and substantially surpassing both \EL{} ($59.0\%$) and Static ($57.5\%$). This indicates that \morph{} benefits from reduced variance and better robustness in smaller networks, narrowing the gap to the upper bound more effectively than in larger-scale settings.

In terms of test loss dynamics, \morph{} consistently follows the trajectory of the fully connected topology across both datasets. Although slightly higher loss values are observed throughout training, the differences remain marginal, and late-stage increases are shared by all methods. This suggests that while the fully connected graph retains a small edge, our approach achieves near-optimal convergence without requiring dense connectivity.

Finally, in Figure~\ref{fig:inv_general} we analyze the inter-node variance of test accuracies, which quantifies the disparity in performance across individual nodes. A higher variance indicates that certain nodes perform substantially worse than others, undermining fairness and overall robustness of the decentralized system. The results show a striking contrast: \EL{} exhibits the highest inter-node variance ($15.50$), revealing severe inconsistency across nodes. In contrast, both the fully connected baseline ($0.018$) and our method \morph{} ($0.013$) achieve almost negligible variance, ensuring nearly uniform accuracy across participants. The static topology yields zero variance by construction, since nodes remain fixed in their communication partners and thus converge to nearly identical models; however, this comes at the cost of significantly lower accuracy (cf. Table~\ref{tab:decentralized-accuracy}). Taken together, these results demonstrate that \morph{} achieves a desirable balance, combining accuracy close to the fully connected upper bound with robustness to inter-node performance disparities, while avoiding the pathological inconsistency of \EL.

\subsection{Impact of Connectivity on Accuracy}
Figure~\ref{fig:cifar_abl_k} shows CIFAR-10 test accuracies with $100$ nodes under connectivity levels $k \in \{3,7,14\}$. As expected, accuracy rises with higher $k$, since nodes access broader neighborhoods. The fully connected baseline achieves $69.3\%$, $70.1\%$, and $69.9\%$, while \morph{} closely follows with $68.9\%$, $69.5\%$, and $69.3\%$, never more than $0.4$ points below the upper bound. This demonstrates that \morph{} preserves strong generalization even at sparse connectivity. 

\EL, however, is highly sensitive: it drops to $60.9\%$ at $k=3$, improving to $65.9\%$ at $k=7$ and $68.0\%$ at $k=14$, but consistently lags behind \morph{} and the baseline. Static shows mixed behavior—only $61.6\%$ at $k=3$, but reaching $69.5\%$ at $k=7$ before falling again to $68.0\%$ at $k=14$, indicating weaker robustness across settings.

\begin{figure}
    \centering
    \includegraphics[width=1.0\linewidth]{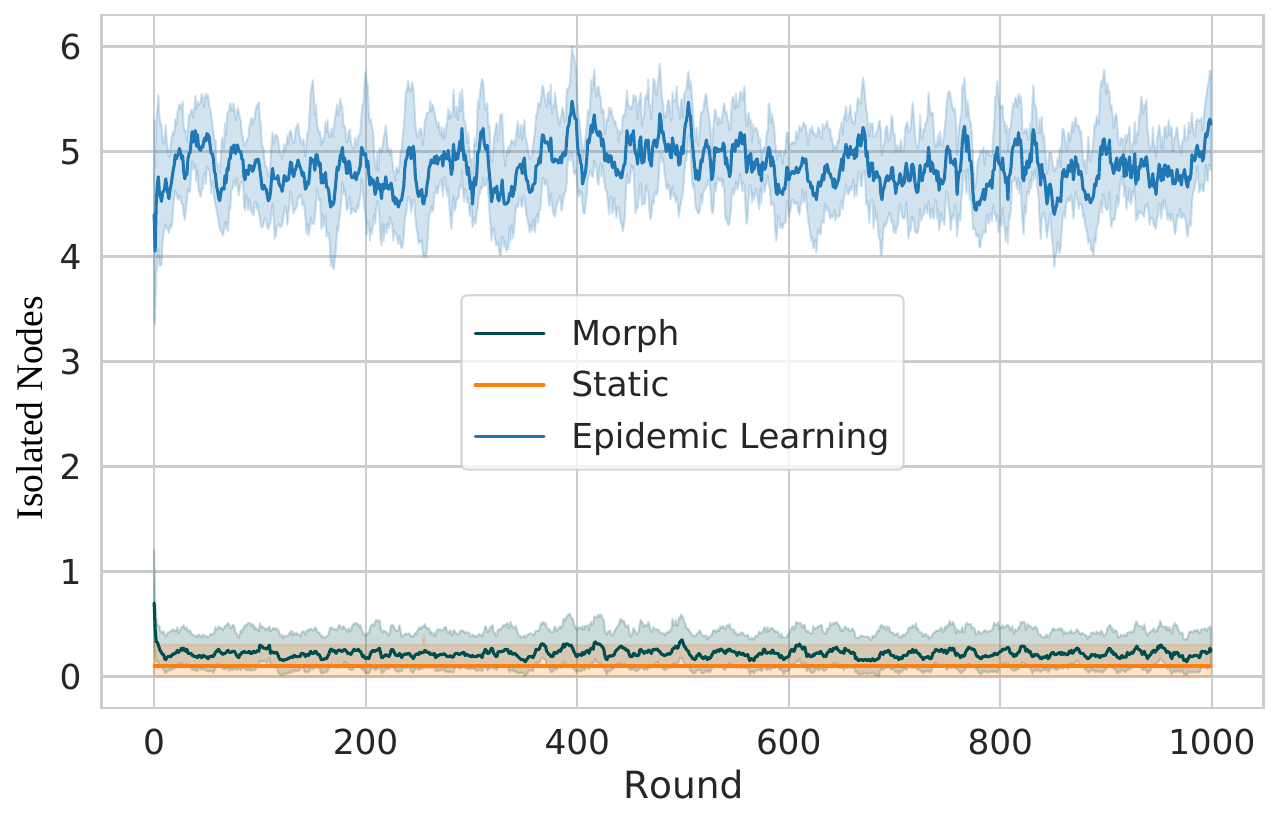}
    \caption{Number of nodes that receive no incoming connection in a network on 100 nodes. These nodes cannot update their model. The random node selection of \EL{} can cause some nodes to become isolated, while \morph{} maintains a connected network.}
    \label{fig:isolated_nodes}
\end{figure}

\begin{figure}[t]
  \centering
  \includegraphics[width=1.0\linewidth]{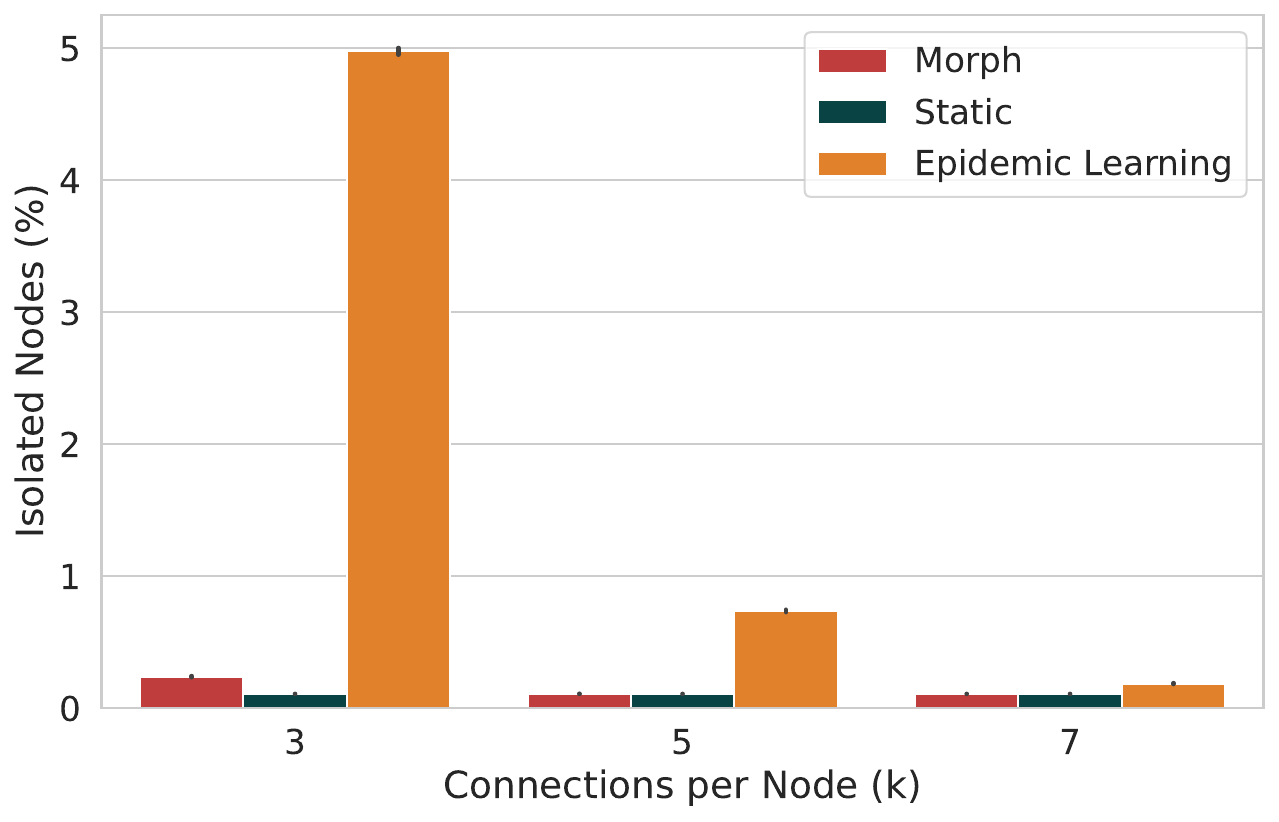}
  \caption{
    Percent of nodes with no incoming connections, isolated nodes, for different algorithms and values of $k$. The plot compares the performance of \EL{} and Static with varying values of $k$ (3, 5, and 7). We observe that a low values of $k$ increases the percentage of isolated nodes in the system when using \EL.
  }
  \label{fig:num_isolated_nodes}
\end{figure}

Connectivity also influences the fraction of isolated nodes in the network. As shown in Figure~\ref{fig:isolated_nodes}, \EL{} consistently produces a subset of nodes that receive no model updates, resulting in isolation. The extent of this isolation strongly depends on the connectivity level $k$ (see Figure~\ref{fig:num_isolated_nodes}). Specifically, \EL{} suffers severe isolation at low connectivity, with an average of $14.1$ isolated nodes at $k=3$, decreasing to $2.0$ at $k=5$ and $0.44$ at $k=7$. This explains its reduced accuracy under sparse topologies. In contrast, \morph{} effectively minimizes isolation, maintaining fewer than one isolated node even at $k=3$. The Static topology trivially avoids isolation ($\approx 0.2$ nodes across all $k$) due to its fixed peer connections, but lacks adaptability to data and topology dynamics. Overall, \morph{} achieves the most favorable balance—preserving robustness under sparse connectivity while maintaining accuracy close to the fully connected upper bound.

\subsection{Impact of Parameters}

\morph{} introduces two key parameters that influence stability and convergence speed: (i) $\beta$, which controls the sharpness of the softmax in Equation~\ref{eq:biased_sampling}, and  (ii) \evalInterval, which defines how frequently nodes compare model similarity with their neighbors.  
Figure~\ref{fig:abl_beta_sim_interval} summarizes their impact. The left panel shows that lower $\beta$ values yield faster and more stable convergence, confirming the importance of biasing neighbor selection through a smoother softmax. The right panel shows that reducing \evalInterval{} below $100$ rounds does not significantly improve accuracy. Since similarity evaluation incurs both communication and computational overhead, larger \evalInterval{} values are generally preferable. However, setting $\Delta_r = 1000$ leads to a noticeable slowdown in convergence, suggesting that overly infrequent updates harm learning.  
In practice, we recommend choosing $\Delta_r < 1000$ to balance efficiency and accuracy. Importantly, the optimal \evalInterval{}depends on system characteristics such as the number of nodes and dataset scale, and thus should be tuned per deployment.

\section{Related Work}
\begin{table*}[t]
\centering
\footnotesize
\setlength{\tabcolsep}{4pt}
\renewcommand{\arraystretch}{1.3}
\setlength{\aboverulesep}{0pt}
\setlength{\belowrulesep}{0pt}
\caption{
Comparison of topology-aware distributed algorithms. A method is \textbf{decentralized} if it requires no central coordinator; \textbf{no global info} means no reliance on node identities or full topology; \textbf{guided adaptation} uses heuristics (not random); \textbf{flexible topology} allows evolving beyond a fixed graph.
}
\label{tab:method-comparison}

\rowcolors{2}{gray!10}{white}
\begin{tabular}{
  p{5cm}cccc
}
\rowcolor{gray!25}
\toprule
\textbf{Method} & \textbf{Decentralized} & \textbf{No Global Info} & \textbf{Guided Adaptation} & \textbf{Flexible Topology} \\
\toprule
Menegatti et al.~\cite{menegattiDynamicTopologyOptimization2024}, Lin et al.~\cite{linReinforcementBasedCommunication2021}, Wang et al.~\cite{wangAcceleratingDecentralizedFederated2023b}, Zhou et al.~\cite{zhouAcceleratingDecentralizedFederated2024}, Tuan et al.~\cite{tuanDFLTopologyOptimization2025} & \xmark & \xmark & \cmark & \cmark \\
Behera et al. (PFedGame)~\cite{beheraPFedGameDecentralizedFederated2024} & \xmark & \xmark & \xmark & \cmark \\
Li et al. (L2C/meta-L2C)~\cite{liLearningCollaborateDecentralized2022} & \cmark & \cmark & \cmark & \xmark \\
Assran et al. (SGP)~\cite{assranStochasticGradientPush2019} & \cmark & \cmark & \xmark & \xmark \\
Song et al. (EquiDyn)~\cite{songCommunicationEfficientTopologiesDecentralized2022} & \cmark & \xmark & \xmark & \xmark \\
De Vos et al. (EL-Local)~\cite{devosEpidemicLearningBoosting2023} & \cmark & \xmark & \xmark & \cmark \\
Bars et al.~\cite{barsRefinedConvergenceTopology2023} & \xmark & \xmark & \cmark & \xmark \\
Dandi et al.~\cite{dandiDataheterogeneityawareMixingDecentralized2022} & \cmark & \cmark & \xmark & \xmark \\
\textbf{\morph{} (this work)} & \cmark & \cmark & \cmark & \cmark \\
\bottomrule
\end{tabular}
\end{table*}

Table~\ref{tab:method-comparison} compares \morph{} to recent topology-aware decentralized learning methods in terms of decentralization, information requirements, adaptation strategy, and topological flexibility. \morph{} is the only protocol that is decentralized, does not require global information, uses guided topology adaptation and adopts a dynamic communication graph.   

\subsection{Fixed Topology Algorithms}
Early work in decentralized learning (DL) typically assumes a fixed communication graph and focuses on improving algorithms or designing static topologies for non-IID data.  
Aketi et al. propose two such methods: NGC~\cite{aketiNeighborhoodGradientClustering2023}, which clusters local and neighbor gradients by similarity, and NGM~\cite{aketiNeighborhoodGradientMean2023}, which averages them for lower overhead.  
Other approaches leverage additional structure: Gao et al.~\cite{gaoGraphNeuralNetwork2022} use a pre-trained GNN to guide aggregation, Esfandiari et al.~\cite{esfandiariCrossGradientAggregationDecentralized2021} introduce Cross-Gradient Aggregation (CGA) via constrained QP, and Song et al.~\cite{songCommunicationEfficientTopologiesDecentralized2022} design EquiStatic, a family of communication-efficient topologies.  
While effective under non-IID data, these methods are ultimately limited by their fixed initial graph.

averaging local and cross-gradients, making it more suitable for bandwidth- or memory-constrained scenarios.

\subsection{Topology-Aware Algorithms with Global Coordination}

Recent methods adapt topologies using global knowledge. Menegatti et al.~\cite{menegattiDynamicTopologyOptimization2024} optimize algebraic connectivity for faster convergence, while Behera et al. (PFedGame)~\cite{beheraPFedGameDecentralizedFederated2024} model aggregation as a cooperative game. Lin et al.~\cite{linReinforcementBasedCommunication2021} use centralized reinforcement learning to optimize peer selection, and Wang et al. (CoCo)~\cite{wangAcceleratingDecentralizedFederated2023b} employ a central solver to jointly assign peers and compression levels.  
Other work, such as Zhou et al.~\cite{zhouAcceleratingDecentralizedFederated2024} and Tuan et al.~\cite{tuanDFLTopologyOptimization2025}, adds edges or predicts topologies to maximize algebraic connectivity.  
While these strategies improve efficiency, they depend on global graph information or central coordination, limiting applicability in fully decentralized settings.

\subsection{Decentralized Dynamic Topology Algorithms}

Fully decentralized methods aim to exploit dynamic topologies without global control. Koloskova et al.~\cite{koloskovaUnifiedTheoryDecentralized2020} provide theoretical guarantees for convergence under time-varying graphs. Li et al.~\cite{liLearningCollaborateDecentralized2022} propose L2C and meta-L2C, which prune dense initial graphs into fixed sparse topologies based on validation loss.  
Assran et al. (SGP)~\cite{assranStochasticGradientPush2019} and Ying et al.~\cite{yingExponentialGraphProvably2021} decompose exponential graphs into dynamic schedules of pairwise exchanges, reducing communication while retaining convergence rates. Song et al. (EquiDyn)~\cite{songCommunicationEfficientTopologiesDecentralized2022} extend this idea by allowing each node to contact one neighbor per round, achieving network-size-independent consensus rates but still bounded by the initial graph.  
De Vos et al. (\EL)~\cite{devosEpidemicLearningBoosting2023} broadcast updates to random peers, improving mixing but lacking guided neighbor selection and assuming global peer knowledge.

\subsection{Peer Dissimilarity as a Topology Signal}
An important open question in decentralized learning is how to select communication partners effectively, especially when data distributions differ significantly across nodes. Recent work has begun to explore data-aware topology construction strategies, highlighting the importance of designing topologies that facilitate information exchange between heterogeneous nodes. Bars et al.~\cite{barsRefinedConvergenceTopology2023} show that communication with dissimilar nodes, those whose local data distributions differ, helps ensure that each node’s neighborhood better approximates the global distribution. Similarly, Dandi et al.~\cite{dandiDataheterogeneityawareMixingDecentralized2022} report that convergence improves when communication weights are aligned with the complementarity of local data, such that nodes with more diverse data distributions are more strongly connected. These findings suggest that, particularly under non-IID conditions, it is advantageous for nodes to communicate with others that have different data characteristics.

\section{Conclusion}
We introduced \morph{}, a fully decentralized learning algorithm that dynamically adapts its communication topology based on local model dissimilarity. By allowing nodes to connect with peers whose models differ meaningfully from their own, \morph{} improves robustness and accelerates convergence under non-IID data distributions. 
Experiments on CIFAR-10 and FEMNIST show that \morph{} consistently outperforms static and epidemic baselines in accuracy, convergence speed, and inter-node variance, while maintaining comparable communication overhead. On CIFAR-10 with 100 nodes, \morph{} achieves a $1.13\times$ improvement over state-of-the-art baselines, and $1.08\times$ on FEMNIST. It also narrows the gap to the fully connected upper bound to within $0.5$ percentage points under sparse connectivity, demonstrating strong adaptability and efficiency.

These findings highlight model dissimilarity as an effective principle for adaptive topology optimization in decentralized learning. 
 
Future work may incorporate additional node-level metrics,such as latency, data diversity, or learning progress, to enhance scalability, fairness, and adaptability in large, dynamic networks.

\printbibliography

\end{document}